%% file: main.tex
\pdfoutput=1
\documentclass[11pt]{article}

\usepackage[final]{acl}

\usepackage{times}
\usepackage{latexsym}

\usepackage[T1]{fontenc}

\usepackage[utf8]{inputenc}

\usepackage{microtype}

\usepackage{inconsolata}

\usepackage{graphicx}

\usepackage[most]{tcolorbox}
\usepackage{enumitem}
\usepackage{booktabs}

\title{Counting Clues:\\A Lightweight Probabilistic Baseline Can Match an LLM}

\author{Furong Jia\thanks{Equal contribution.} \\
  Duke University\\
  \texttt{flora.jia@duke.edu} \\\And
  Yuan Pu\footnotemark[1] \\
   Duke University\\
  \texttt{yuan.pu@duke.edu} \\\AND
  Finn Guo \\
   Duke University\\
  \texttt{xiaofeng.guo@duke.edu}\And
  Monica Agrawal \\
   Duke University\\
  \texttt{monica.agrawal@duke.edu}}

\newcommand{\modelName}{Frequency-Based Probabilistic Ranker}
\newcommand{\modelNameAbbrev}{FBPR}

\begin{document}
\maketitle
\begin{abstract}
Large language models (LLMs) excel on multiple-choice clinical diagnosis benchmarks, yet it is unclear how much of this performance reflects underlying probabilistic reasoning. We study this through questions from MedQA that ask for the most likely diagnosis. We introduce \modelName{} (\modelNameAbbrev), a lightweight method that scores options with a smoothed Naïve Bayes over concept-diagnosis co-occurrence statistics from a large corpus. When co-occurrence statistics are sourced from Dolma, the pretraining corpus for OLMo, and RedPajama, an open reproduction of the LLaMA pretraining-data recipe, \modelNameAbbrev{} achieves comparable performance to the corresponding LLMs. Furthermore, direct LLM inference and \modelNameAbbrev{} exhibit complementary strengths, correctly answering largely different subsets of questions with overlap barely above chance. Combining them with a simple hybrid method yields a notable gain over either method alone.
These findings show that explicit probabilistic baselines can account for a substantial portion of the benchmark performance on diagnosis questions, while serving as historically grounded reference points and complementary signals for hybridization.
\end{abstract}

\section{Introduction}
\label{sec:introduction}

Large language models (LLMs) have been used for tasks requiring probabilistic reasoning, including direct inference and estimation of priors \citep{paruchuri2024odds, feng2024bird, nafar2025extracting}.
However, our understanding of probabilistic reasoning in LLMs is still nascent. In this work, we study the probabilistic reasoning required to choose the most likely medical diagnosis given a clinical scenario.
LLMs have achieved high performance across medical benchmarks, leading to speculation that they possess advanced clinical reasoning capabilities \citep{Kung2022PerformanceOCA, Singhal2022LargeLMA, Singhal2025TowardEMA, Nori2023CapabilitiesOGA}. This narrative is reinforced by claims of LLMs achieving diagnostic performance on complex cases that surpass human experts \citep{mcduff2025towards, nori2025sequential}. However, a growing number of works also reveal potential deficiencies in the clinical reasoning and decision-making of current LLMs \citep{chen2024clinicalbench}. For example, expert systems such as DxPlain still surpass LLMs on unpublished diagnostic cases \citep{barnett1987dxplain, berner1994performance, feldman2025dedicated}. This suggests a possible disconnect between performance on medical licensing exams and true probabilistic reasoning, with research highlighting LLM failures in metacognition \citep{griot2025large}, processing new information under uncertainty \citep{mccoy2025language}, and aligning with physicians' reasoning in medical question-answering \citep{hao2025medpair}.

\begin{figure*}[htbp]
    \centering
    \includegraphics[width=1\linewidth]{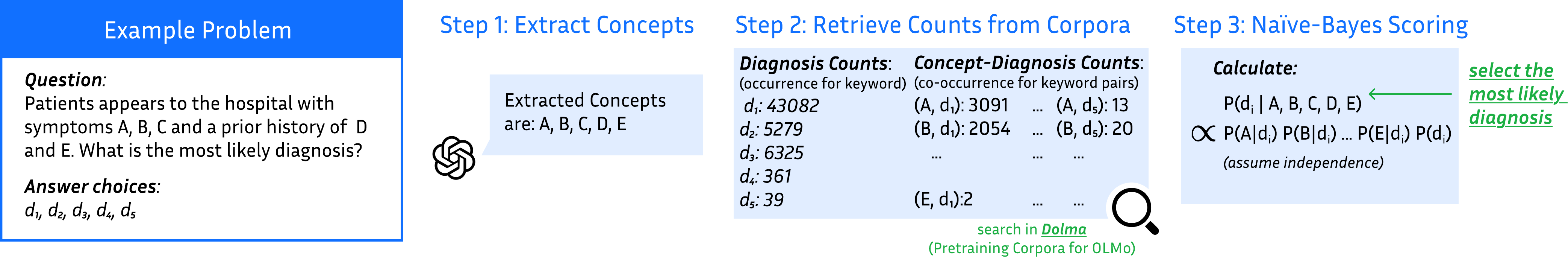}
    \caption{\modelName: (1) concept extraction from the question, (2) corpus frequency retrieval, and (3) compute a Naive Bayes score to select the most likely diagnosis.}
    \label{fig:pipeline}
\end{figure*}

Building on these observations, a key evaluation gap is clarifying how much reported accuracy on medical QA benchmarks reflects exploitation of corpus‑level co‑occurrence statistics versus more structured clinical probabilistic reasoning. LLMs' clinical knowledge has been linked to how frequently relevant clinical information appears in their pretraining corpora \citep{jia2025diagnosing}. Widely used biomedical QA benchmarks are predominantly knowledge‑heavy (67.2\% of questions across 11 datasets), with models performing worse on the reasoning‑heavy subsets \citep{thapa2025disentangling}. Complementary synthetic evidence shows that LLMs can achieve 64\% on a fictional benchmark about an invented organ (Glianorex), while physicians reach only 27\% \citep{griot2025pattern}, indicating that strong scores can arise from linguistic cues and test-taking heuristics over domain knowledge. These patterns suggest revisiting earlier paradigms that leveraged statistical association, such as earlier clinical decision-support systems (CDSS) like DxPlain \citep{barnett1987dxplain, berner1994performance}. DxPlain combines computer-interpretable clinical guidelines with an expert-designed, data-refinable Bayesian network to provide probabilistic, patient-specific decision support. Recent evidence suggests that such systems can still surpass LLMs on unpublished diagnostic cases \citep{nee2010clinical, elkin2010introduction, bauer2002internal, feldman2025dedicated}.

Recent work uses LLM-derived probabilities for Bayesian reasoning, including Bayesian-network inference and parameterization \citep{feng2024bird, nafar2025extracting}. In contrast, we ask how much a coarser, low-complexity, and transparent, frequency‑based method can achieve on diagnosis questions using only an LLM's pretraining corpus.
Therefore, we introduce \textbf{\modelName{} (\modelNameAbbrev)} that eschews complex reasoning (Figure \ref{fig:pipeline}), and models diagnosis using a Naive Bayes-like probability derived from the co-occurrence frequency of diagnosis and clinical concepts from large-scale corpora, Dolma \citep{soldaini-etal-2024-dolma} and RedPajama \citep{weber2024redpajama}. Dolma is the pretraining corpus for OLMo, whereas RedPajama is an open reproduction of the LLaMA pretraining data constructed following the recipe documented in the LLaMA technical report. We evaluate \modelNameAbbrev{} on a diagnostic subset of \textsc{MedQA}-USMLE, which consists of United States Medical Licensing Examination-style multiple-choice questions \citep{jin2021disease}. We also evaluate OLMo Instruct 7B \citep{groeneveld-etal-2024-olmo}, an open-weight instruction-tuned LLM pretrained on the Dolma corpus, and LLaMA 65B \citep{touvron2023llama}, pretrained on the RedPajama corpus, for performance comparison with \modelNameAbbrev{} applied to their respective pretraining corpora. We found that Dolma-based \modelNameAbbrev{} achieved an accuracy (46.7\%) substantially above random chance (20\%) and closely matched OLMo Instruct (44.1\%). Similar results were obtained from \modelNameAbbrev{} based on RedPajama (44.5\%) and LLaMA 65B (47.0\%). A simple exploratory hybrid method combining \modelNameAbbrev{} with LLMs further improves performance (47.3\% and 50.8\%) across both corpora, highlighting the complementarity of the two approaches and the potential of using them in combination. As an additional evaluation, we observe a similar pattern on a diagnosis subset of \textsc{MedMCQA} \citep{pal2022medmcqa}, where Dolma-based \modelNameAbbrev{} achieves 52.7\% accuracy compared with 54.9\% for OLMo Instruct 7B (Appendix~\ref{app:diagnosis_subset}).

\section{Methodology}

\subsection{Data and Task Formulation}
\label{sec:task}
We evaluate two corpus-model pairings: (1) OLMo Instruct 7B (\texttt{OLMo-7B-0724-Instruct}) with its pretraining corpus, Dolma, and (2) LLaMA 65B with RedPajama, an open reproduction of the LLaMA pretraining data. OLMo Instruct 7B is an instruction-tuned model, while LLaMA 65B is substantially larger and can reliably produce answers in the required format under zero-shot prompting (Appendix~\ref{app:LLM_eval}). This transparency enables a fair, corpus-aligned comparison between our statistical method and the LLM baselines.

We focus on a diagnosis subset of the \textsc{MedQA} benchmark ($n = 719$; details in Appendix~\ref{app:diagnosis_subset}), which is not among the documented data source of Dolma, instruction-tuning dataset for OLMo Instruct, or RedPajama. Each question consists of a patient scenario followed by a prompt for the most likely diagnosis and five candidate diagnoses $D=\{d_1,\ldots,d_5\}$. The goal is to predict the most likely diagnosis $d^\star\in D$ from the scenario description. As an additional evaluation, we also construct a diagnosis subset of \textsc{MedMCQA} \citep{pal2022medmcqa}, a dataset drawn from Indian medical entrance exams. The subset consists of 639 multiple-choice questions with four candidate answers per question; we use the same task formulation with $D=\{d_1,\ldots,d_4\}$ (Appendix~\ref{app:diagnosis_subset}).

Out of these 719 \textsc{MedQA} questions, 256 are from the Step 1 domain of the US medical licensing exam, which focuses on basic science knowledge, and 463 are from the Step 2\&3, which emphasize clinical reasoning and patient management. It is widely recognized that, while Step 1 emphasizes factual recall, Steps 2\&3 require an application of knowledge through clinical reasoning and therefore demand stronger reasoning ability \citep{jin2021disease}.

\subsection{\modelName}
\label{sec:method}

Our pipeline proceeds through the three stages depicted in Fig.~\ref{fig:pipeline} and detailed below:

\paragraph{Stage 1: concept extraction}
\label{sec:keywords}
For each question, we prompt GPT-4o (\texttt{gpt-4o-2024-11-20}) to extract $k$ concise (each $\le$ 4 words) clinical concepts $\mathbf{x}=\{x_1,\dots,x_k\}$ that were emphasized in the presented clinical scenario. The prompt omits any mention of diagnosis to avoid bias toward selectively ``diagnostic'' features and instead aims to obtain an objective set of symptoms mentioned. We further ask GPT-4o to label each extracted concept with whether it is mentioned positively or negated.  GPT-4o is used only for extraction and polarity labeling, not for scoring answer choices. To assess the extent to which downstream performance depends on the capability of the concept extractor, we additionally replicate the pipeline using the open-weight Qwen2.5-32B-Instruct \citep{qwen2.5,qwen2} as an alternative extractor, which has lower reported performance than GPT-4o on \textsc{MedQA}. Prompts and examples are detailed in Appendix~\ref{app:concept_extraction} and Appendix~\ref{app:examples}.

\paragraph{Stage 2: corpus frequency retrieval}
\label{sec:counts}
We leverage \textit{Infini-gram} \citep{Liu2024InfiniGram}, an efficient tool for querying token and phrase frequencies in massive corpora, to obtain from Dolma or RedPajama
\begin{itemize}[leftmargin=*,nosep]
    \item the number of occurrences of candidate diagnosis $d$ as $C(d)$, and
    \item the number of co-occurrences of candidate diagnosis $d$ along with extracted concept $x_i$ as $C(d, x_i)$.
\end{itemize}
We also expand each candidate diagnosis and extracted concept using variants in spacing and capitalization to account for the fact that \textit{Infini-gram} performs tokenization before searching. All co-occurrences are counted within \textit{Infini-gram}'s maximum AND window of 1000 tokens. More details are in Appendix~\ref{app:infinigram}.

\paragraph{Stage 3: Naive-Bayes scoring}
\label{sec:bayes}
To isolate corpus-frequency signal rather than model all clinical dependencies, we assume concept independence and conditional independence given each candidate diagnosis, estimate the posterior $P(d\mid x_1,\dots,x_k)$ for each candidate diagnosis, and choose the one with the highest probability given the extracted concepts.

Let $d\in D$ and abbreviate $\mathbf{x}=\{x_1,\dots,x_k\}$.

\begin{align}
P(d\mid\mathbf{x})
&= \frac{P(\mathbf{x}\mid d)\, P(d)}{P(\mathbf{x})} \label{eq:b1}\\
&= \frac{\left[\prod_{i=1}^{k'} P(x_i\mid d)\right]\,P(d)}
         {P(\mathbf{x})} \label{eq:b2}\\
&\approx \frac{\left[\prod_{i=1}^{k'} \frac{C'(d,x_i)}{C(d)}\right] \frac{C(d)}{N}}{P(\mathbf{x})}\label{eq:b3}\\
&= \frac{C(d)^{-(k'-1)}}{N\cdot P(\mathbf{x})} \left[\prod_{i=1}^{k'} C'(d,x_i)\right] .\label{eq:b4}
\end{align}

Step~\eqref{eq:b2} is based on the naive conditional-independence
assumption. Step~\eqref{eq:b3} substitutes empirical estimates
$P(x_i \mid d) \approx C'(d,x_i)/C(d)$ and
$P(d) \approx C(d)/N$, where $N$ is the total token count of the
pre-training corpus.

Given the affirmation/negation label, we took $C'(d, x_i) = C(d, x_i)$ for every (candidate diagnosis, affirmed concept) pair and excluded the negated concepts from the score calculation. $k'$ thus equaled the number of affirmed concepts in each question.

The denominator $N \cdot P(\mathbf{x})$ does not depend on $d$ and is
dropped in the calculation of the score for comparing candidate diagnoses. With add-$\delta$ smoothing ($\delta = 0.001$) and a logarithm, we have the score for each candidate diagnosis as

\resizebox{0.95\columnwidth}{!}{$
\hspace{-10pt}
\begin{aligned}
S_d=\sum_{i=1}^{k'} \log\left(C'(d,x_i)+\delta\right)
 -(k'-1)\log\left(C(d)+\delta\right)
\end{aligned}
$}
and the answer by \modelNameAbbrev{} is the candidate diagnosis with the highest score within each question:
\begin{align*}
    \hat{d}=\text{argmax}_{d\in D} S_d.
\end{align*}

\section{Results}
\label{sec:results}

\subsection{\modelNameAbbrev{} Performance}

\begin{table}[htbp]
\centering
\begin{tabular}[width=1\linewidth]{cc}
\toprule
Method & Accuracy \\
\midrule
Random Chance & 20\% \\
\midrule
\modelNameAbbrev{} \scriptsize{on Dolma} & 46.7\%\\
OLMo Instruct 7B & 44.1\% \\
Hybrid \scriptsize{(OLMo + Dolma)} & 47.3\%\\
\midrule
\modelNameAbbrev{} \scriptsize{on RedPajama} & 44.5\%\\
LLaMA 65B & 47.0\% \\
Hybrid \scriptsize{(LLaMA 65B + RedPajama)} & 50.8\% \\
\bottomrule
\end{tabular}
\caption{Prediction accuracy on five-option \textsc{MedQA} diagnosis subset ($n$ = 719).}
\label{tab:methods_accuracy}
\end{table}

\paragraph{\textit{k}=5}
We fixed $k=5$ a priori and prompted GPT-4o to extract five concepts from each diagnostic question and assign each concept an affirmation or negation label.
16 of the 719 questions contained negated concepts (see Appendix~\ref{app:concept_extraction} for details).

\begin{figure}[htbp]
    \centering
    \includegraphics[width=0.9\linewidth]{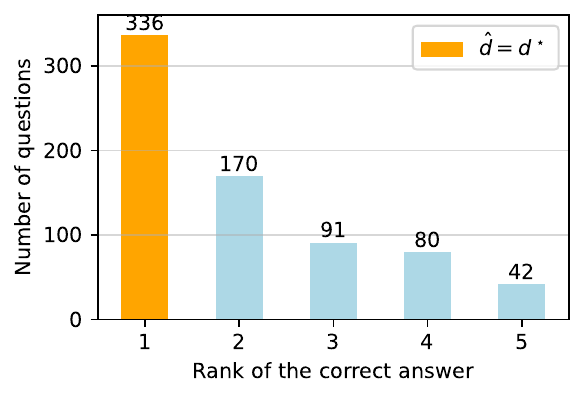}
    \caption{Distribution of the rank of the correct diagnosis of each question by Dolma-based \modelNameAbbrev{} with $k=5$ on \textsc{MedQA} diagnosis subset.}
    \label{fig:dolma_ignoreNegated_correctOption_ranks}
\end{figure}

\begin{figure}[b!]
    \centering
    \includegraphics[width=\linewidth]{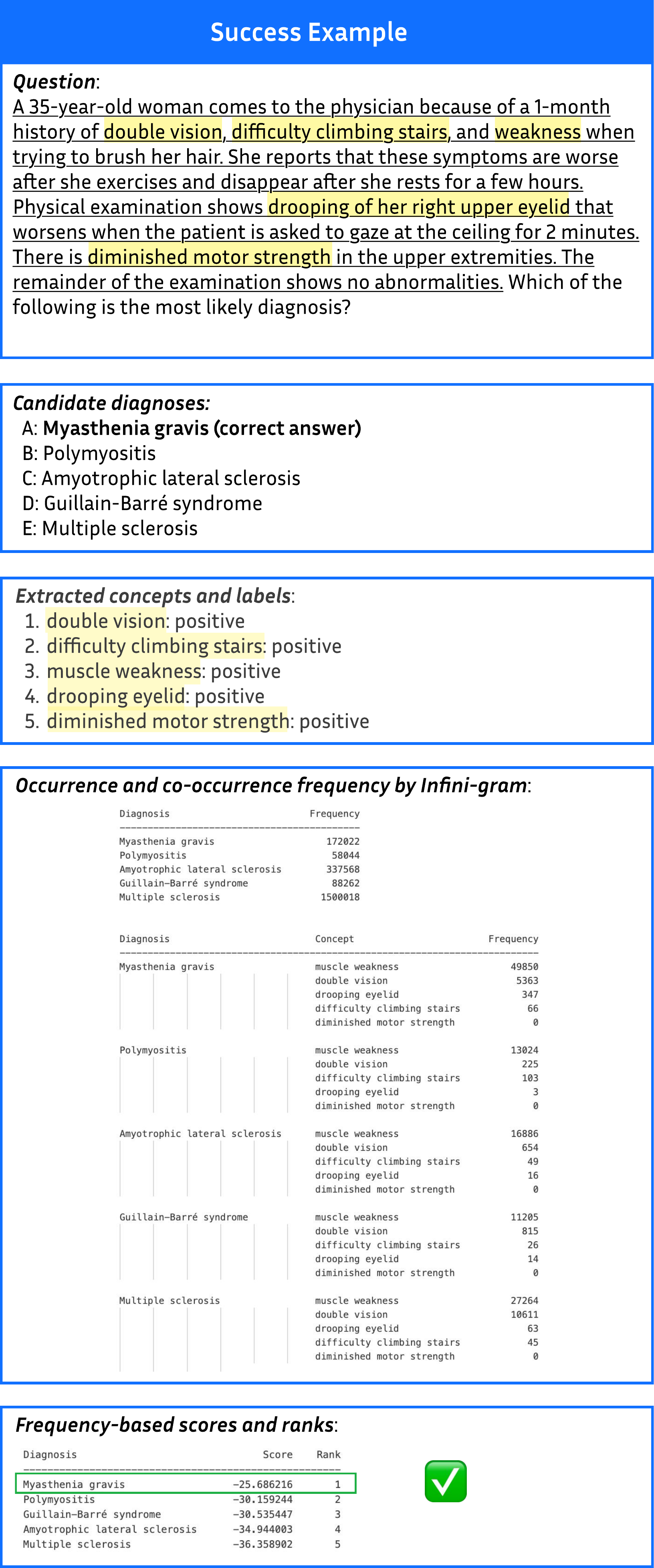}
    \caption{A successful example of \modelNameAbbrev{} on a \textsc{MedQA} diagnosis question. The diagnosis occurrence and (diagnosis, concept) co-occurrence frequencies are retrieved by \textit{Infini-gram} from \texttt{v4\_rpj\_llama\_s4} index, i.e., from the RedPajama corpus. By \textit{Infini-gram} design, some of these frequencies are estimates.}
    \label{fig:success_example}
\end{figure}

With $k$=5, Dolma-based \modelNameAbbrev{} reached 46.7\% accuracy and the RedPajama-based one 44.5\% (Table~\ref{tab:methods_accuracy}). The correct diagnosis is most frequently assigned the highest rank by \modelNameAbbrev{}, with progressively fewer instances receiving ranks 2 through 5 (Figure~\ref{fig:dolma_ignoreNegated_correctOption_ranks}). The observed monotonic pattern shows that this simple frequency-based method appears to capture clinically meaningful links between presenting concepts and diagnoses. The same pattern holds for RedPajama-based \modelNameAbbrev{} (Figure~\ref{fig:rpj_ignoreNegated_correctOption_ranks}). Figure~\ref{fig:success_example} shows an example from \textsc{MedQA} where \modelNameAbbrev{} arrives at a correct prediction in this setting.

Dolma-based \modelNameAbbrev{} correctly identified the most likely diagnosis for 51.2\% of Step 1 questions but only 44.3\% of Step 2\&3 questions (Table~\ref{tab:dolmaANDrpj_accuracy_MedQAtrain} in Appendix). This difference likely reflects the heavier multi-step reasoning demands of Step 2\&3, where \modelNameAbbrev{}'s occurrence frequency-based reasoning is less effective. However, the gap is modest, suggesting that even simple frequency-based Bayesian strategies can account for a notable portion of performance often attributed to complex reasoning.

We further explored the relationship between \modelNameAbbrev's internal certainty and its actual correctness by using the top-1 scores softmaxed within each question as a proxy for certainty (Figure~\ref{fig:dolma_ignoreNegated_topScoreSoftmax_dist} in Appendix). For Dolma-based \modelNameAbbrev, correctly predicted questions have higher softmaxed top scores compared to incorrectly predicted ones (median 0.94 vs. 0.86). 27.4\% of questions have a softmaxed highest score exceeding 0.99; the accuracy of \modelNameAbbrev{} on these questions is 59.4\%, higher than the general 46.7\% achieved on the full dataset.

\paragraph{Unrestricted \textit{k}}
We also experimented with prompting GPT-4o without a restriction on $k$ (see Appendix~\ref{app:concept_extraction} for details). Under this setting, \modelNameAbbrev{} achieved 45.1\% on Dolma and 40.9\% on RedPajama. Its rank distribution follows the same decreasing pattern as under $k=5$ (Figure~\ref{fig:rpj_ignoreNegated_correctOption_ranks} in Appendix) and has a slightly higher performance on Step 2\&3 than Step 1 (Table~\ref{tab:dolmaANDrpj_accuracy_MedQAtrain}).

\paragraph{Alternative concept extractor}
To better ensure reproducibility and robustness, we next examine how downstream performance changes when replacing GPT-4o with the open-weight Qwen2.5-32B-Instruct for concept extraction. With $k=5$, \modelNameAbbrev{} achieves 44.0\% accuracy using Qwen2.5-32B-Instruct, compared with 46.7\% using GPT-4o. With unrestricted $k$, the corresponding accuracies are 43.4\% and 45.1\%, respectively (Table~\ref{tab:dolmaANDrpj_accuracy_MedQAtrain}). Overall, \modelNameAbbrev{} achieves similar performance and remains comparable to the corpus-aligned OLMo Instruct 7B (44.1\%) under the alternative concept extractor.

\paragraph{Additional evaluation on \textsc{MedMCQA}}
On the \textsc{MedMCQA} diagnosis subset, when using the open-weight Qwen2.5-32B-Instruct as the concept extractor, Dolma-based \modelNameAbbrev{} achieves 52.7\% accuracy, compared with 54.9\% for OLMo Instruct 7B, showing a similar performance pattern to that observed on \textsc{MedQA} (Appendix~\ref{app:diagnosis_subset}).

\subsection{Alignment with LLM Performance}
To assess alignment with a model trained on the same underlying distribution, we compare the Dolma-based \modelNameAbbrev{} to OLMo Instruct 7B (pretrained on the Dolma corpus) and the RedPajama-based \modelNameAbbrev{} to LLaMA 65B, both under a zero-shot setting (details in Appendix~\ref{app:LLM_eval}).

Dolma-based \modelNameAbbrev{} and OLMo Instruct are both correct on 24.8\% of questions. This observed joint accuracy is slightly higher than the expected joint success of around $44.1\%\times 46.7\% \approx 20.6\%$ under an independence assumption, indicating only a mild correlation. \modelNameAbbrev{} alone is correct on 21.9\% of questions, and OLMo alone on 19.3\%.

Across the diagnostic subset, the two methods agree on 38.3\% of the questions (shown as the diagonal in Figure~\ref{fig:olmo_confusion_matrix}). 64.7\% of these agreed answers are correct. This indicates that inter-method agreement is a helpful reliability cue. This pattern also holds for RedPajama-based \modelNameAbbrev{} and LLaMA 65B (Appendix~\ref{app:rpj_llama}).

\subsection{An Exploratory Hybrid Method}
We use a simple hybrid method to examine whether \modelNameAbbrev{} and direct LLM inference provide complementary predictions. The hybrid returns the LLM's answer when it appears among \modelNameAbbrev{}'s top $m$ options and otherwise returns \modelNameAbbrev{}'s top-ranked option. We use $m = \lceil |D|/2 \rceil = 3$, as it is the median ranking for the five-option task. We report the full $m$ sweep in Appendix~\ref{app:K_sweep} to demonstrate the overall improvement across different $m$ values.
The hybrid method improves accuracy over both component methods in each comparison (Table~\ref{tab:methods_accuracy}). The gains are statistically significant for LLaMA 65B with RedPajama but not for OLMo with Dolma (Appendix~\ref{app:stats}).

\section{Discussion}
\label{sec:discussion}

Our simple frequency-based concept–diagnosis co‑occurrence method attains accuracy on \textsc{MedQA} and \textsc{MedMCQA} diagnostic questions comparable to LLMs pre-trained on the same corpora, indicating that surface distributional priors can account for a substantial portion of the benchmark performance on the diagnosis questions studied. Despite similar accuracy, the LLM appears to exploit different mechanisms over the same corpus and thus offers complementary signals that combine constructively,  as our simple hybrid demonstrates.

The strength of this signal is particularly notable given the minimal preprocessing in our implementation. \modelNameAbbrev{} performs no synonym or semantic normalization (e.g., ``high blood pressure,'' ``elevated blood pressure,'' and ``hypertension'' remain distinct concepts).  While incorporating richer concept normalization could further improve \modelNameAbbrev{} performance (see Appendix~\ref{synonym} for illustrative cases), the simplicity of our method actually underscores how even a very naive baseline can be competitive. Additionally, future work could explore additional methods of probing LLMs to distinguish corpus-based pattern retrieval from more structured forms of clinical probabilistic reasoning.

\newpage
\section*{Limitations}
Our analysis targets diagnosis-focused subsets of \textsc{MedQA} and \textsc{MedMCQA}, and the concept set is extracted by an external LLM (GPT-4o, or Qwen2.5-32B-Instruct as an alternative), making results sensitive to prompting and extraction choices. Frequency estimates are obtained via \textit{Infini-gram} \citep{Liu2024InfiniGram}, which returns occurrence-level (rather than document-level) counts and approximates very frequent strings; precise document-level co-occurrence is deferred due to resource constraints. We further limit evaluation to two transparent corpus–model pairs: (1) Dolma and OLMo Instruct 7B, (2) RedPajama and LLaMA 65B. Neither is state-of-the-art on this benchmark and thus may not reflect the frontier of LLM performance. Nonetheless, our aim is not to introduce a high-performing system but to isolate how far an explicit, low-complexity probabilistic baseline can go using pretraining corpus statistics.

\section*{Ethics Statement}
In writing this paper, we used an AI assistant to correct grammatical errors. During the coding process, we utilized AI tools for code completion.

\newpage
\bibliography{custom}

\appendix

\input{appendix}

\end{document}

%% file: appendix.tex
\appendix
\onecolumn

\section{Diagnosis subset formation}\label{app:diagnosis_subset}
\paragraph{\textsc{MedQA}} The \textsc{MedQA} benchmark is a collection of United States Medical Licensing Examination (USMLE)-style questions. Each question stem consists of a clinical vignette describing a patient scenario, followed by a query that typically requires identifying the most likely diagnosis, the appropriate next step in management, or a related clinical decision. We used the version where each question provides five candidate answer options, exactly one of which is correct.

We focus on diagnosis-related questions in the \textsc{MedQA} training set. We first identified questions in which the query (the last sentence of the question stem) explicitly contains the word ``diagnosis''. After converting all query text to lowercase, we found and kept questions with the two most common phrasings: ``which of the following is the most likely diagnosis?'' (588 questions) and ``what is the most likely diagnosis?'' (131 questions). This query sentence in each question was stripped before feeding the clinical scenario into GPT-4o for concept extraction and labeling.

\paragraph{\textsc{MedMCQA}}
\label{app:medmcqa}
In addition to \textsc{MedQA}, we constructed a diagnosis subset of \textsc{MedMCQA} \citep{pal2022medmcqa}, a multiple-choice dataset built from Indian medical entrance exams with four candidate answers per question. From the \textsc{MedMCQA} training set, we kept single-answer questions whose stem asks for the ``most likely diagnosis'' (642 questions) and removed three questions whose answer options are images, resulting in $n=639$. As with \textsc{MedQA}, the trailing query was stripped from each stem before concept extraction. As shown in Table~\ref{tab:medmcqa}, Dolma-based \modelNameAbbrev{} also performs comparably to OLMo Instruct 7B on \textsc{MedMCQA}.

\begin{table}[htbp]
\centering
\begin{tabular}{lc}
\toprule
Method & Accuracy \\
\midrule
Random chance & 25.0\% \\
\modelNameAbbrev{} on Dolma ($k=5$) & 52.7\% \\
\modelNameAbbrev{} on Dolma (unrestricted $k$) & 52.4\% \\
OLMo Instruct 7B & 54.9\% \\
\bottomrule
\end{tabular}
\caption{Prediction accuracy on the four-option \textsc{MedMCQA} diagnosis subset ($n=639$). Concepts for \modelNameAbbrev{} are extracted by Qwen2.5-32B-Instruct (Appendix~\ref{app:qwen_extractor}), and OLMo Instruct 7B was evaluated with the zero-shot prompt in Appendix~\ref{app:LLM_eval}}
\label{tab:medmcqa}
\end{table}

\section{Concept extraction and labeling}
\label{app:concept_extraction}

We use GPT-4o only to extract surface clinical concepts from the vignette; the subsequent ranking stage does not use LLM inference over answer choices and relies only on corpus counts. For both concept extraction and affirmation/negation labeling, we queried the GPT-4o API (\texttt{gpt-4o-2024-11-20}) with \texttt{temperature} set to 0.0 and \texttt{top\_logprobs} set to 0, ensuring mostly deterministic outputs. Example extracted concepts and labels can be found in Appendix~\ref{app:examples}.

\subsection{\textit{k}=5} We prompted GPT-4o for concept extraction with a system prompt
\begin{tcolorbox}[breakable, enhanced]
You are a medical expert that extracts the five most informative clinical concepts from each passage.
\end{tcolorbox}
and a query prompt that encodes both the number of concepts to extract (\texttt{n}=5) and the question stem (\texttt{cleaned\_question}), excluding the trailing query such as ``What is the most likely diagnosis?''.
\begin{tcolorbox}[breakable, enhanced]
Given the passage below, list exactly \{\texttt{n}\} canonical **clinical** concepts.\\
\\
Rules for each concept:\\
• Must be clinically relevant (symptom, physical sign, diagnosis, or treatment).\\
• Prefer the passage’s wording **if** it is already $\leq$ 4 words; otherwise shorten or normalize to a standard term $\leq$ 4 words (no numbers or strength grades).\\
• Skip generic exam words and any term containing the stems: exam, test, lab, imaging, manage, work-up.\\
• The term itself must not contain a comma.\\
\\
Output formatting: lower-case, comma-separated list of exactly \{\texttt{n}\} concepts, with no other text.\\
\\
Passage:\\
\{\texttt{cleaned\_question}\}
\end{tcolorbox}

Then the extracted concepts (\texttt{keywords}) along with the cleaned question stem (\texttt{question\_text}) are sent to GPT-4o for affirmation/negation labeling in a query prompt
\begin{tcolorbox}[breakable, enhanced]
Question:\\
{\texttt{question\_text}}\\
\\
Keywords (comma-separated, KEEP ORDER):\\
\texttt{', '.join(keywords)}\\
\\
Return the single required line exactly as specified.
\end{tcolorbox}
alongside a system prompt
\begin{tcolorbox}[breakable, enhanced]
You are a precise medical NLP assistant tagging keyword polarity for MedQA-style prompts.\\

Task:\\
Given a MedQA question text and an ordered, comma-separated list of keywords, decide for EACH keyword whether it is affirmed/present/tested (``positive'') or explicitly denied/absent/ruled-out (``negative'') based ONLY on the question text.\\
\\
Rules:\\
- ``positive'' if the question affirms/treats the keyword as present/relevant/tested or as a correct/true option.\\
- ``negative'' ONLY if the question explicitly negates or rules it out (e.g., ``NOT'', ``EXCEPT'', ``absent'', ``no evidence of'', ``which is NOT'').\\
- If merely mentioned (no explicit negation), label ``positive''.\\
- Be careful with NOT/EXCEPT questions: items asked for as NOT/EXCEPT are ``negative''.\\
- Judge strictly from the given text; do not add outside knowledge.\\
\\
Output FORMAT (STRICT):\\
Return a SINGLE LINE string containing EXACTLY one item per keyword in the SAME ORDER, separated by comma+space, \\
each item in the form ``$<$keyword$>$: positive'' or ``$<$keyword$>$: negative''. \\
Do NOT add any extra text, quotes, explanations, or newlines.
\end{tcolorbox}

16 out of 719 \textsc{MedQA} diagnosis subset questions contain negated concepts (15 with a single negation and 1 with two).

\subsection{No restriction on \textit{k}}
\begin{figure}[ht]
    \centering
    \includegraphics[width=0.7\linewidth]{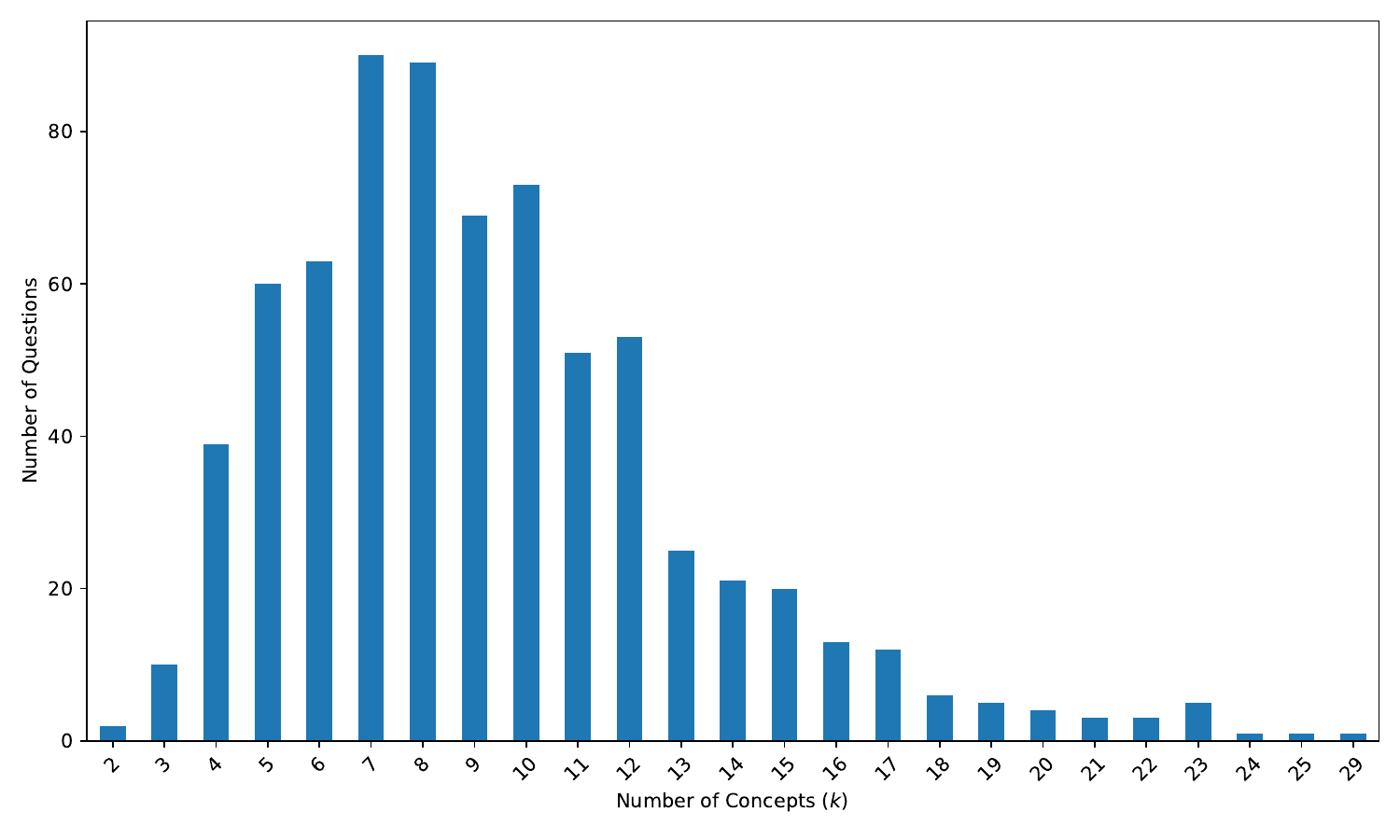}
    \caption{Distribution of the number of concepts per question extracted by GPT-4o when prompt contains no restriction.}
    \label{fig:kNoRestrict_kdist}
\end{figure}

We also explored prompting GPT-4o without restricting $k$ for concept extraction. Specifically, ``five'' in the concept extraction system prompt and ``exactly \{\texttt{n}\}'' in the query prompt were omitted. GPT-4o returned a varying $k$ across questions with a median (IQR) of 9 (7, 11) (Figure~\ref{fig:kNoRestrict_kdist}). 44 questions had negated concepts (30 with a single negated concept, 6 with two, 7 with three, and 1 with five).

\subsection{Concept extractor choice}
\label{app:extractor_comparison}

We use GPT-4o for phrase-level concept extraction because our downstream corpus query relies on exact-match co-occurrence counts. Off-the-shelf clinical entity extractors are often optimized for ontology mapping and may fragment coherent clinical phrases into generic entities, which are poorly matched to this retrieval setting. Examples comparing entities extracted using Google Healthcare API and phrase-level concepts from GPT-4o are demonstrated in Table~\ref{tab:extractor_comparison}. For example, in one MedQA vignette, GPT-4o extracts phrase-level concepts such as ``difficulty climbing stairs'' and ``weakness brushing hair,'' while the Google Healthcare API extracts more fragmented entities such as ``weakness'' and ``hair.'' Searching for co-occurrences with generic fragments such as ``hair'' can introduce substantial noise because the clinical context has been lost.

\begin{table}[htbp]
\centering
\begin{tabular}{p{0.28\linewidth}p{0.32\linewidth}p{0.32\linewidth}}
\toprule
Text summary & GPT-4o phrase-level concepts & Google Healthcare API entities \\
\midrule
Patient with double vision, difficulty climbing stairs, and weakness when brushing hair &
double vision; difficulty climbing stairs; weakness brushing hair; symptoms worse after exercise; symptoms improve with rest; drooping right upper eyelid &
double vision; weakness; hair; symptoms; worse; drooping; eyelid; diminished; motor strength; upper extremities \\
\midrule
Patient with Crohn disease on natalizumab with difficulty walking and ataxia &
crohn disease; difficulty walking; gait ataxia; limb ataxia; natalizumab &
Crohn disease; difficulty walking; Crohn disease; natalizumab; intestinal symptoms; severe; unresponsive; therapies; gait; limb; ataxia \\
\bottomrule
\end{tabular}
\caption{Examples illustrating why phrase-level extraction is better matched to exact-match co-occurrence retrieval than fragmented clinical entity extraction.}
\label{tab:extractor_comparison}
\end{table}

\label{app:qwen_extractor}
We also used an alternative LLM for concept extraction to assess the sensitivity of \modelNameAbbrev{} to the concept extractor. We ran concept extraction with Qwen2.5-32B-Instruct, an open-weight, general-purpose LLM that is less accurate than GPT-4o on \textsc{MedQA}. We used the same extraction prompts as for GPT-4o, for both $k=5$ and unrestricted $k$, with greedy decoding (temperature = 0). Without a restriction on $k$, Qwen2.5-32B-Instruct returned a median of 8 concepts per question. With Qwen2.5-32B-Instruct as the extractor, \modelNameAbbrev{} on Dolma reaches 44.0\% ($k=5$) and 43.4\% (unrestricted $k$) for \textsc{MedQA}, remaining comparable to OLMo Instruct 7B (44.1\%) (Table~\ref{tab:dolmaANDrpj_accuracy_MedQAtrain}).

\begin{table*}[htbp]
\centering
\begin{tabular}{llcccc}
\toprule
Extractor & Corpus & $k$ & all & Step 1 & Step 2\&3\\
\midrule
GPT-4o & Dolma & 5 & 46.7\% & 51.2\% & 44.3\%  \\
GPT-4o & Dolma & unrestricted & 45.1\% & 47.7\% & 43.6\%  \\
GPT-4o & RedPajama & 5 & 44.5\% & 46.1\% & 43.6\%  \\
GPT-4o & RedPajama & unrestricted & 40.9\% & 40.2\% & 41.3\%  \\
\midrule
Qwen2.5-32B-Instruct & Dolma & 5 & 44.0\% & 46.1\% & 42.8\% \\
Qwen2.5-32B-Instruct & Dolma & unrestricted & 43.4\% & 47.3\% & 41.3\% \\
\bottomrule
\end{tabular}
\caption{Accuracy of \modelNameAbbrev{} on \textsc{MedQA} diagnosis subset ($n$=719), based on occurrence frequency from different corpora and with two settings of $k$, stratified on the medical licensing exam domain. Concepts are extracted by GPT-4o (main setting) or by Qwen2.5-32B-Instruct (Appendix~\ref{app:qwen_extractor}).}
\label{tab:dolmaANDrpj_accuracy_MedQAtrain}
\end{table*}

\section{\textit{Infini-gram} usage}
\label{app:infinigram}

\textit{Infini-gram} provides a large-scale n-gram index that enables fast retrieval of occurrence and co-occurrence statistics from web-scale corpora. It performs search at the token level, converting input words or phrases into tokens before counting their occurrences. Variations such as the presence of a leading space and lowercase versus capitalized forms influence how tokens are generated and therefore affect the returned counts. Our interest, however, lies in the frequency of candidate diagnoses or (candidate diagnosis, concept) pairs in a case-agnostic and position-agnostic manner. To address this, we generated four surface-form variations for each candidate diagnosis or concept (original, leading space, all lowercase, and first-letter capitalized) and combined them using an OR clause, i.e.,
\begin{align*}
&\texttt{text\_original} \text{ OR } \\
    &\texttt{text\_original\_leading\_space} \text{ OR } \\
    & \texttt{text\_all\_lowercase} \text{ OR } \\
    & \texttt{text\_all\_lowercase\_leading\_space} \text{ OR } \\
    &\texttt{text\_capitalized} \text{ OR } \\
    &\texttt{text\_capitalized\_leading\_space}.
\end{align*}
OR clauses are supported by \textit{Infini-gram} to obtain aggregated occurrence counts for the OR connected components, and thus help return our desired occurrence frequency of candidate diagnoses.

To obtain co-occurrence of a candidate diagnosis and a concept, we feed \textit{Infini-gram} with a clause connecting their OR clauses with an AND operator, i.e.,
\begin{align*}
    (\\
    &\texttt{diagnosis\_original} \text{ OR } \\
    &\texttt{diagnosis\_original\_leading\_space} \text{ OR } \\
    & \texttt{diagnosis\_all\_lowercase} \text{ OR } \\
    & \texttt{diagnosis\_all\_lowercase\_leading\_space} \text{ OR } \\
    &\texttt{diagnosis\_capitalized} \text{ OR } \\
    &\texttt{diagnosis\_capitalized\_leading\_space}\\
    ) &\text{ AND } (\\
    &\texttt{concept\_original} \text{ OR } \\
    &\texttt{concept\_original\_leading\_space} \text{ OR } \\
    &\texttt{concept\_all\_lowercase} \text{ OR } \\
    &\texttt{concept\_all\_lowercase\_leading\_space} \text{ OR } \\
    &\texttt{concept\_capitalized} \text{ OR } \\
    &\texttt{concept\_capitalized\_leading\_space}
    ).
\end{align*}
Note that we de-duplicate the text variations before connecting them in OR clause. For most words or phrases, there are four unique variants: all-lowercase with and without a leading space, and initial-capitalized with and without a leading space. Some text, for example the ones with internal capitalization, have more than four variations created by our algorithm. A related complication is that \textit{Infini-gram} supports OR clauses with at most four components, which requires splitting the query into multiple OR clauses or OR–AND clauses. For the former, a summation of individual OR clause count is sufficient for the overall count; for the latter, we apply inclusion–exclusion to compute the counts.

By \textit{Infini-gram} design, the occurrences of clauses connected by AND can only be examined within a window of 1000 tokens at maximum. When any clause connected by AND has an occurrence frequency larger than 500000, the returned co-occurrence frequency is an estimate. All co-occurrence counts in this paper use this maximum window of 1000 tokens.

We used two indices provided by \textit{Infini-gram}:
\begin{itemize}
    \item \texttt{v4\_dolma-v1\_7\_llama}: Dolma-v1.7 corpus tokenized by LLaMA-2 \citep{touvron2023llama2} consists of 3,403,336,408 documents and 2,604,642,372,173 tokens.
    \item \texttt{v4\_rpj\_llama\_s4}: RedPajama corpus tokenized by LLaMA-2 consists of 931,361,530 documents and 1,385,942,948,192 tokens.
\end{itemize}

\section{Large language models evaluation}
\label{app:LLM_eval}

We performed zero-shot prompting to get LLM prediction of most-likely diagnosis, using the Hugging Face checkpoints \texttt{allenai/OLMo-7B-0724-Instruct-hf} and \texttt{huggyllama/llama-65b}. Below is an example query for LLM:
\begin{tcolorbox}[breakable, enhanced]
\textbf{Question:} A 70-year-old man comes to the physician for the evaluation of an 8-week history of blood in his stool. Two months ago, he had an episode of bronchitis and was treated with amoxicillin. Since then, he has noticed blood in his stool and on the toilet paper occasionally. The patient has had intermittent constipation for the past 5 years. Six months ago, he had severe left lower quadrant pain and fever that resolved with antibiotic therapy. He underwent a colonoscopy 3 years ago, which did not show any evidence of malignancy. He takes levothyroxine for hypothyroidism. He had smoked one pack of cigarettes daily for 45 years, but quit smoking 10 years ago. He drinks one glass of red wine every night. He appears pale. He is 180 cm (5 ft 11 in) tall and weighs 98 kg (216 lb); BMI is 32 kg/m2. His temperature is 36°C (96.8°F), pulse is 85/min, and blood pressure is 135/80 mm Hg. Physical examination shows pale conjunctivae. Cardiopulmonary examination shows no abnormalities. The abdomen is soft and nontender with no organomegaly. Digital rectal examination shows no masses. Test of the stool for occult blood is positive. Laboratory studies show:\\
Hemoglobin 11 g/dL\\
Mean corpuscular volume 76 $\mu$m$^3$\\
Red cell distribution width 17\% (N = 13–15)\\
Leukocyte count 5,000/mm$^3$\\
Which of the following is the most likely diagnosis?"\\
\\
\textbf{Options:}\\
A: Colorectal carcinoma\\
B: Diverticulosis\\
C: Ischemic colitis\\
D: Hemorrhoids\\
E: Pseudomembranous colitis\\
\\
\textbf{Answer: }
\end{tcolorbox}
OLMo Instruct 7B generated the following output in response to the query above:
\begin{tcolorbox}[breakable, enhanced]
B: Diverticulosis\\
\\
Explanation:\\
\\
The patient's history of intermittent constipation, positive occult blood test, and the absence of any evidence of malignancy on previous colonoscopy make diverticulosis the most likely diagnosis.
\end{tcolorbox}
An extractor was created to extract the answer from the LLM response. In the example above, the extracted answer refers to option choice B. We adapt a method similar to \citet{jeong2024medical} and apply five tiers in order:
\begin{enumerate}
    \item \textbf{Stand-alone letter regex.} A strict regex finds exact, stand-alone option letters (A--E). If exactly one letter---or several copies of the same letter---appears, it is the prediction.
    \item \textbf{LaTeX/Markdown emphasised letter.} Patterns such as \texttt{\textbackslash strong\{C\}}, \texttt{**B**}, \texttt{\_\_C\_\_} are recognised; the wrapped letter is the prediction.
    \item \textbf{Numbered-list dominant letter.} Patterns of the form \texttt{"1.B 2.B 3.B ..."} are parsed; if a single letter dominates the list (count $\ge 2$), it is the prediction.
    \item \textbf{Single-option phrase match.} The text is normalised; a prediction is made only if one option's phrase appears while all others are absent.
    \item \textbf{Most-frequent stand-alone letter.} If multiple distinct letters appear without consensus from tiers 1--4, the most-frequent stand-alone letter is the prediction.
\end{enumerate}

\section{Alternative scoring settings}
In addition to the scoring method presented in the main text (i.e., \textbf{ignore negated concepts}), we experimented with two additional scoring formulations:
\begin{enumerate}
    \item \textbf{Affirmation/negation agnostic}: affirmation and negation labels of concepts are disregarded, so $C'(d, x_i) = C(d, x_i)$ for every candidate diagnosis–concept pair, and $k' = k$.
    \item \textbf{Reward absence of negated concepts}: for a negated concept $x_i$ we consider the number of times the candidate diagnosis $d$ does \textit{not} co-occur with $x_i$, so
    \begin{align*}
        C'(d, x_i) =
        \begin{cases}
            C(d, x_i) &\text{if $x_i$ is affirmed} \\
            C(d) - C(d, x_i) &\text{if $x_i$ is negated}, \\
        \end{cases}
    \end{align*}
    with $k'=k$. Note that non-co-occurrence within an \textit{Infini-gram} window does not imply clinical absence: the concept may be expressed through a synonym, fall outside the window, or appear elsewhere in the document. This rule may therefore convert retrieval failures into negative evidence.
\end{enumerate}
The three scoring methods yield highly consistent results on \textsc{MedQA} diagnosis subset, largely due to the small number of negated concepts.

\section{Synonym normalization}
\label{synonym}
\modelNameAbbrev{} uses exact matching between extracted concepts and answer options and applies no synonym expansion or semantic normalization. This appendix illustrates the practical implications of the choice through three representative cases. We focus on cases where the correct answer has zero co-occurrence with at least one extracted concept: in such cases an extracted concept contributes no signal to the correct option, while synonym expansion could in principle introduce signal. For each such concept, we obtain three synonym candidates from GPT-4o (temperature = 0) and use infini-gram with inclusion–exclusion to compute the exact disjunctive count of the original concept together with its three synonyms against each answer option; scores are then recomputed with these expanded counts.

\paragraph{Example 1: Lack of synonyms negatively impacts the score.}
For the question below, the extracted concept \textit{throbbing headache} has zero co-occurrences with the correct answer \textit{Hypertensive emergency} under exact matching, while the synonym set \{\textit{pulsatile headache, pounding headache, vascular headache}\} is able to retrieve co-occurrences. Including these forms is sufficient to move \textit{Hypertensive emergency} from rank 4 to rank 1.

\begin{tcolorbox}[breakable, enhanced]
\textbf{Question:} A 72-year-old woman comes to the emergency department 4 hours after the sudden onset of a diffuse, dull, \underline{throbbing headache}. During this time, she also reports blurred vision, nausea, and one episode of vomiting. She has a history of hypertension and type 2 diabetes mellitus. Her medications include hydrochlorothiazide, lisinopril, atorvastatin, and metformin. She has smoked 1 pack of cigarettes daily for the past 30 years and drinks 1–2 glasses of wine per day. Her temperature is 36.6 °C (97.9 °F), pulse is 90/min, respirations are 14/min, and blood pressure is 185/110 mm Hg. Fundoscopic examination shows bilateral blurring of the optic disc margins. Physical and neurologic examinations show no other abnormalities. A complete blood count and serum concentrations of electrolytes, urea nitrogen, and creatinine are within the reference range. Urinalysis shows 2+ protein but no WBCs or RBCs. Which of the following is the most likely diagnosis?

\textbf{Options:}\\
A: Ischemic stroke\\
B: Subarachnoid hemorrhage\\
C: Idiopathic intracranial hypertension\\
\underline{D: Hypertensive emergency}\\
E: Transient ischemic attack\\
\end{tcolorbox}

\paragraph{Example 2: canonical synonyms do not necessarily change the score.}
For the question below, the extracted concept \textit{hypertension} has a clear near-canonical paraphrase in \textit{high blood pressure}, along with variants such as \textit{arterial hypertension} and \textit{elevated blood pressure}. Adding these forms does not change the rank of the correct answer: the correct option continues to receive zero co-occurrence with all three synonyms, and the score gap to competitors widens as other options obtain higher co-occurrence counts with the synonym expansion.

\begin{tcolorbox}[breakable, enhanced]
\textbf{Question:} A 65-year-old man is brought into the emergency department by his wife for slurred speech and right-sided weakness. The patient has a significant past medical history of \underline{hypertension} and hyperlipidemia. The wife reports her husband went to bed last night normally but woke up this morning with the symptoms mentioned. Physical examination shows right-sided hemiparesis along with the loss of vibration and proprioception. Cranial nerve examination shows a deviated tongue to the left. What is the most likely diagnosis?

\textbf{Options:}\\
A: Medial pontine syndrome\\
B: Lateral pontine syndrome\\
\underline{C: Dejerine syndrome}\\
D: Wallenberg syndrome\\
E: Weber syndrome
\end{tcolorbox}

\paragraph{Example 3: some concepts have no canonical normalization target.}
For the question below, the extracted concept \textit{burnt rubber smell} has zero co-occurrences across all five options. Unlike the clean synonym pair \textit{hypertension} / \textit{high blood pressure}, this concept has no canonical medical paraphrase. Normalization here forces an arbitrary choice between lay-description preservation and information-discarding abstraction, with no surface-form rule to select between them.

\begin{tcolorbox}[breakable, enhanced]
\textbf{Question:} A 50-year-old man presents to his primary care provider complaining of smelling abnormal odors on several occasions. He says that he \underline{smells burnt rubber} even though there is nothing burning and no one around him can smell what he does. This symptom has been intermittently bothering him for the past 6 months. Also during this period, he had occasional nosebleeds. He works as a high school teacher. Although his work gets a little stressful around the exam season, he says he is able to cope well. Family history is unremarkable. He does not smoke or drink alcohol and denies the use of any medication. Physical examination reveals unilateral nasal obstruction with some dried blood in the nasal passage. What is the most likely diagnosis?

\textbf{Options:}\\
A: Psychomotor epilepsy\\
\underline{B: Neuroblastoma}\\
C: Hypnagogic hallucination\\
D: Substance abuse\\
E: Schizophrenia
\end{tcolorbox}

These examples suggest that synonym normalization is potentially useful but not mechanically straightforward. Expansions to synonyms can recover missing signal in some cases, yet it may leave the correct option unchanged or even strengthen competing options in others. Moreover, some clinical descriptions lack a clear canonical synonym. We therefore view semantic normalization as a promising future extension rather than a simple pre-processing step. At the same time, the fact that \modelNameAbbrev{} remains competitive under this deliberately minimal and naive exact-match setup suggests that raw corpus-level frequency signals already capture substantial diagnostic information.

\section{Corpus-model alignment and hybrid analyses}

\subsection{RedPajama-based \modelNameAbbrev{} and LLaMA 65B alignment}
\label{app:rpj_llama}
For RedPajama-based \modelNameAbbrev{} and LLaMA 65B, their joint accuracy is 24.2\% on the \textsc{MedQA} training diagnosis set; LLaMA alone was correct on 22.8\% of questions, while \modelNameAbbrev{} alone has an accuracy of 20.3\%. They gave the same answer to 34.5\% of the cases, of which 70.2\% are correct.

\subsection{Hybrid method variants}
\label{app:confgate}
\begin{figure}[htbp]
    \centering
    \includegraphics[width=0.9\linewidth]{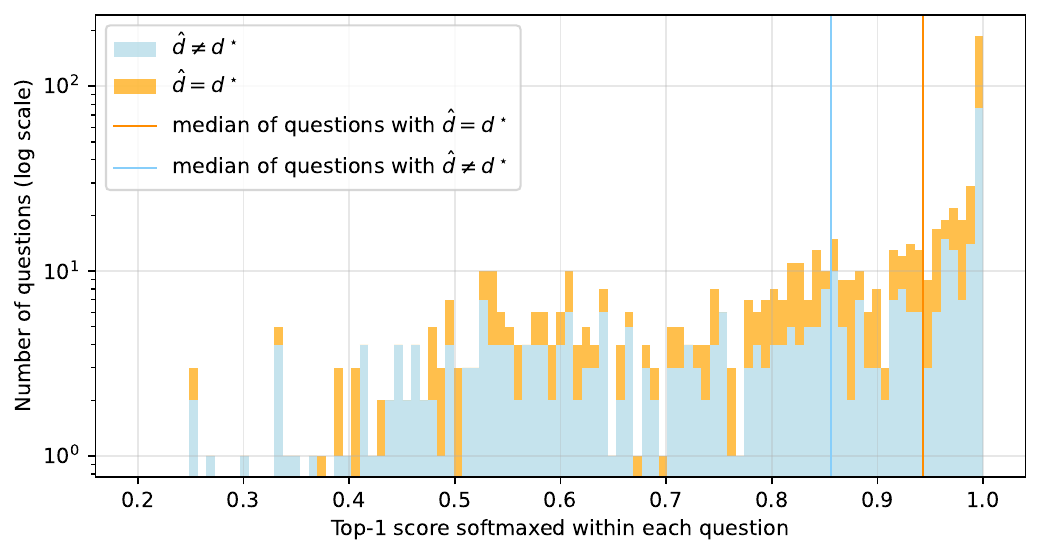}
    \caption{Stacked histogram of per-question softmaxed top-1 scores calculated by Dolma-based \modelNameAbbrev{} with $k=5$. A larger softmaxed top-1 score indicates that \modelNameAbbrev{}'s chosen diagnosis is much more strongly favored by frequency-based Naive Bayes relative to the other candidates. }
    \label{fig:dolma_ignoreNegated_topScoreSoftmax_dist}
\end{figure}

We compare three hybrid rules that combine the LLM regex letter with \modelNameAbbrev{}'s per-option log-scores $\{S_d\}_{d \in D}$. Let
\[
  c \;=\; \max_{d \in D} \frac{\exp S_d}{\sum_{d' \in D} \exp S_{d'}}
\]
denote the within-question softmax of \modelNameAbbrev{}'s top-1 score (its distribution on the Dolma pair is shown in Figure~\ref{fig:dolma_ignoreNegated_topScoreSoftmax_dist}): a value near $1$ indicates \modelNameAbbrev{} places almost all probability mass on a single option (high certainty); a value near $1/|D| = 0.2$ indicates the five options score similarly (no useful signal).

\paragraph{Rank gate.}
We use \modelNameAbbrev{}'s corpus signal as coarse-grained filtering. We adopt the LLM's prediction only when \modelNameAbbrev{} deems it reasonable. Specifically, if the LLM's letter falls within \modelNameAbbrev{}'s top-$m$ ranked options, we choose the LLM's output. Otherwise, \modelNameAbbrev{} vetoes the LLM and we fall back to \modelNameAbbrev{}'s top-1 choice. The cutoff $m$ can be any integer in $\{1, \ldots, |D|\}$; the main paper reports results at $m$ $= \lceil |D|/2 \rceil = 3$ (the median rank for $|D| = 5$), a choice that requires no hyperparameter tuning. A full sweep over $m$ $\in \{1, \ldots, 5\}$ is reported in \S\ref{app:K_sweep}.

\paragraph{Confidence gate.}
Return \modelNameAbbrev{}'s top-1 when $c > \tau$; otherwise return the LLM's letter. Trust \modelNameAbbrev{} only when its softmax mass concentrates on a single option; default to the LLM when \modelNameAbbrev{} is uncertain. This introduces one threshold, $\tau$.

\paragraph{Combined gate.}
Return \modelNameAbbrev{}'s top-1 when $c > \tau$; otherwise, apply the rank gate. Combines the strengths of the two: trust \modelNameAbbrev{} unconditionally when it is highly confident; fall back to the rank-veto logic everywhere else.

\begin{table}[htbp]
\centering
\small
\setlength{\tabcolsep}{5pt}
\begin{tabular}{lccc|ccc}
\toprule
& \multicolumn{3}{c|}{LLaMA 65B + RedPajama} & \multicolumn{3}{c}{OLMo Instruct + Dolma} \\
Method & Acc & $\Delta_{\text{LLM}}$ & $\Delta_{\text{FBPR}}$ & Acc & $\Delta_{\text{LLM}}$ & $\Delta_{\text{FBPR}}$ \\
\midrule
LLM       & 47.0          & ---  & ---  & 44.1          & ---  & ---  \\
\modelNameAbbrev{} alone & 44.5          & ---  & ---  & 46.7          & ---  & ---  \\
\midrule
\textbf{Rank gate (main)}     & \textbf{50.8} & +3.7 & +6.3 & 47.3          & +3.2 & +0.6 \\
\midrule
Confidence gate, $\tau{=}0.99$ & 48.4          & +1.4 & +3.9 & 49.2          & +5.2 & +2.5 \\
Confidence gate, $\tau{=}0.95$ & 47.4          & +0.4 & +2.9 & 48.8          & +4.7 & +2.1 \\
Confidence gate, $\tau{=}0.90$ & 47.6          & +0.6 & +3.1 & 48.8          & +4.7 & +2.1 \\
\midrule
Combined gate, $\tau{=}0.99$  & 49.5          & +2.5 & +5.0 & \textbf{50.1} & +6.0 & +3.4 \\
Combined gate, $\tau{=}0.95$  & 48.4          & +1.4 & +3.9 & 49.5          & +5.4 & +2.8 \\
Combined gate, $\tau{=}0.90$  & 48.3          & +1.3 & +3.8 & 49.4          & +5.3 & +2.7 \\
\bottomrule
\end{tabular}
\caption{Accuracy (\%) and accuracy change relative to the LLM baseline ($\Delta_{\text{LLM}}$) and to \modelNameAbbrev{}-alone baseline ($\Delta_{\text{FBPR}}$) for the three hybrid variants on both corpus-model pairs at three confidence thresholds $\tau$. Bold accuracy marks each pair's strongest configuration.}
\label{tab:confgate}
\end{table}

\paragraph{Analysis.}
The best variant depends on which baseline is stronger a priori. On LLaMA 65B + RedPajama (LLM > \modelNameAbbrev{}), the rank gate is strongest ($50.8\%$): it effectively uses \modelNameAbbrev{} as a coarse-level filter while relying on the LLM for fine-grained selection. On OLMo + Dolma (\modelNameAbbrev{} > LLM), trusting \modelNameAbbrev{}'s confidence as a stronger primary signal is superior, making the combined gate at $\tau=0.99$ the best ($50.1\%$). Accuracy remains stable across $\tau \in \{0.90, 0.95, 0.99\}$. We retain the rank gate as our main result because it is simple and yields consistent gains across both pairs.

\subsection{Rank-gate cutoff $m$ ablation}
\label{app:K_sweep}
The rank-gate cutoff $m$ controls how lenient \modelNameAbbrev{} is with the LLM: the LLM letter is kept if it falls within \modelNameAbbrev{}'s top-$m$ ranked options. The main paper uses $m$ $= \lceil |D|/2 \rceil = 3$ in the five-option setting. Table~\ref{tab:K_sweep} reports the rank gate at every $m$ $\in \{1, 2, 3, 4, 5\}$. $m$$=1$ keeps the LLM only when it agrees with \modelNameAbbrev{}'s top-1, so the rule degenerates to \modelNameAbbrev{} alone. $m$$=5$ keeps the LLM unconditionally (since it always lies within \modelNameAbbrev{}'s top-5), so the rule degenerates to LLM alone, except on the $32$ questions where the LLaMA 65B model returns \textit{None} to be extracted by the regex rules, so they fall back to \modelNameAbbrev{} top-1. In general, all $m$ values for meaningful hybrid ($m$ $\in \{2, 3, 4\}$) would lead to improvement over the single method.

\begin{table}[htbp]
\centering
\small
\setlength{\tabcolsep}{5pt}
\begin{tabular}{cccc|ccc}
\toprule
& \multicolumn{3}{c|}{LLaMA 65B + RedPajama} & \multicolumn{3}{c}{OLMo Instruct + Dolma} \\
$m$ & Acc & $\Delta_{\text{LLM}}$ & $\Delta_{\text{FBPR}}$ & Acc & $\Delta_{\text{LLM}}$ & $\Delta_{\text{FBPR}}$ \\
\midrule
1                  & 44.5          & $-2.5$          & $\phantom{+}0.0$ & 46.7          & $+2.6$ & $\phantom{+}0.0$ \\
2                  & 48.3          & $+1.3$          & $+3.8$           & \textbf{49.4} & $+5.3$ & $+2.6$ \\
\textbf{3 (main)}  & \textbf{50.8} & $+3.7$          & $+6.3$           & 47.3          & $+3.2$ & $+0.6$ \\
4                  & 50.5          & $+3.5$          & $+6.0$           & 46.7          & $+2.6$ & $\phantom{+}0.0$ \\
5                  & 48.8          & $+1.8$          & $+4.3$           & 44.1          & $\phantom{+}0.0$ & $-2.6$ \\
\bottomrule
\end{tabular}
\caption{Rank-gate accuracy (\%) and change relative to the LLM baseline ($\Delta_{\text{LLM}}$) and \modelNameAbbrev{}-alone baseline ($\Delta_{\text{FBPR}}$) for $m$ $\in \{1, \ldots, 5\}$. $m$ $= \lceil |D|/2 \rceil = 3$ (paper main) is the peak on LLaMA + RedPajama; $m$ $= 2$ is the peak on OLMo + Dolma. The corpus-dependent best $m$ mirrors the direction asymmetry of the confidence threshold $\tau$: when the LLM is the stronger baseline, the larger $m$ wins; when \modelNameAbbrev{} is stronger, the smaller $m$ wins. The boundary cases $m$ $= 1$ (= \modelNameAbbrev{} alone, $\Delta_{\text{FBPR}} = 0$) and $m$ $= 5$ ($\approx$ LLM alone, $\Delta_{\text{LLM}} \approx 0$) provide reference.}
\label{tab:K_sweep}
\end{table}

\subsection{Why combining helps on one pair but not the other: a bin-stratified view}
\label{app:bins}
To understand mechanistically how the combined gate's accuracy differs from its two component rules, we stratify the 719 questions into the two bins induced by \modelNameAbbrev{}'s certainty $c$ at $\tau = 0.99$. In each bin, the combined gate is deterministically identical to one component:
\begin{itemize}
    \item When $c > \tau$ (high certainty): combined gate $=$ confidence gate $=$ \modelNameAbbrev{}'s top-1.
    \item When $c \le \tau$ (low certainty): combined gate $=$ rank gate.
\end{itemize}
Table~\ref{tab:bins} reports each method's accuracy within each bin.

\begin{table}[htbp]
\centering
\small
\setlength{\tabcolsep}{4pt}
\begin{tabular}{lcc|cc}
\toprule
& \multicolumn{2}{c|}{LLaMA 65B + RedPajama} & \multicolumn{2}{c}{OLMo + Dolma} \\
Method                    & $c > \tau$ ($n{=}228$) & $c \le \tau$ ($n{=}491$) & $c > \tau$ ($n{=}197$) & $c \le \tau$ ($n{=}522$) \\
\midrule
LLM regex                 & 49.6          & 45.8          & 40.6          & 45.4          \\
\modelNameAbbrev{} alone  & 50.0          & 42.0          & 59.4          & 42.0          \\
\midrule
Rank gate                 & \textbf{54.0} & \textbf{49.3} & 49.2          & \textbf{46.6} \\
Confidence gate           & 50.0          & 47.7          & \textbf{59.4} & 45.4          \\
Combined gate             & 50.0          & \textbf{49.3} & \textbf{59.4} & \textbf{46.6} \\
\bottomrule
\end{tabular}
\caption{Per-bin accuracy (\%) at $\tau = 0.99$. By construction, the combined gate matches the confidence gate in the $c > \tau$ bin and the rank gate in the $c \le \tau$ bin. Bold marks the strongest method per bin/pair.}
\label{tab:bins}
\end{table}

\paragraph{Corpus dependency of the combined gate.}
The relative success of the combined gate is determined by its performance across two confidence bins. In the high-certainty bin ($c > \tau$), strictly adopting \modelNameAbbrev{}'s top-1 prediction improves accuracy on OLMo but degrades it on LLaMA compared to the rank gate. In the low-certainty bin ($c \le \tau$), however, the rank gate uniformly outperforms the confidence gate on both pairs, as its fallback mechanism recovers correct predictions even when \modelNameAbbrev{} is uncertain. Consequently, the combined gate achieves an overall gain on OLMo where both conditions align, but falls slightly behind the rank gate on LLaMA due to the high-certainty degradation. This demonstrates that the combined gate's effectiveness is an inherent property of the corpus, specifically the reliability of \modelNameAbbrev{}'s highly confident predictions.

\section{Additional results and examples}

\subsection{Confusion matrix}
\begin{figure}[htbp]
    \centering
    \includegraphics[width=0.5\linewidth]{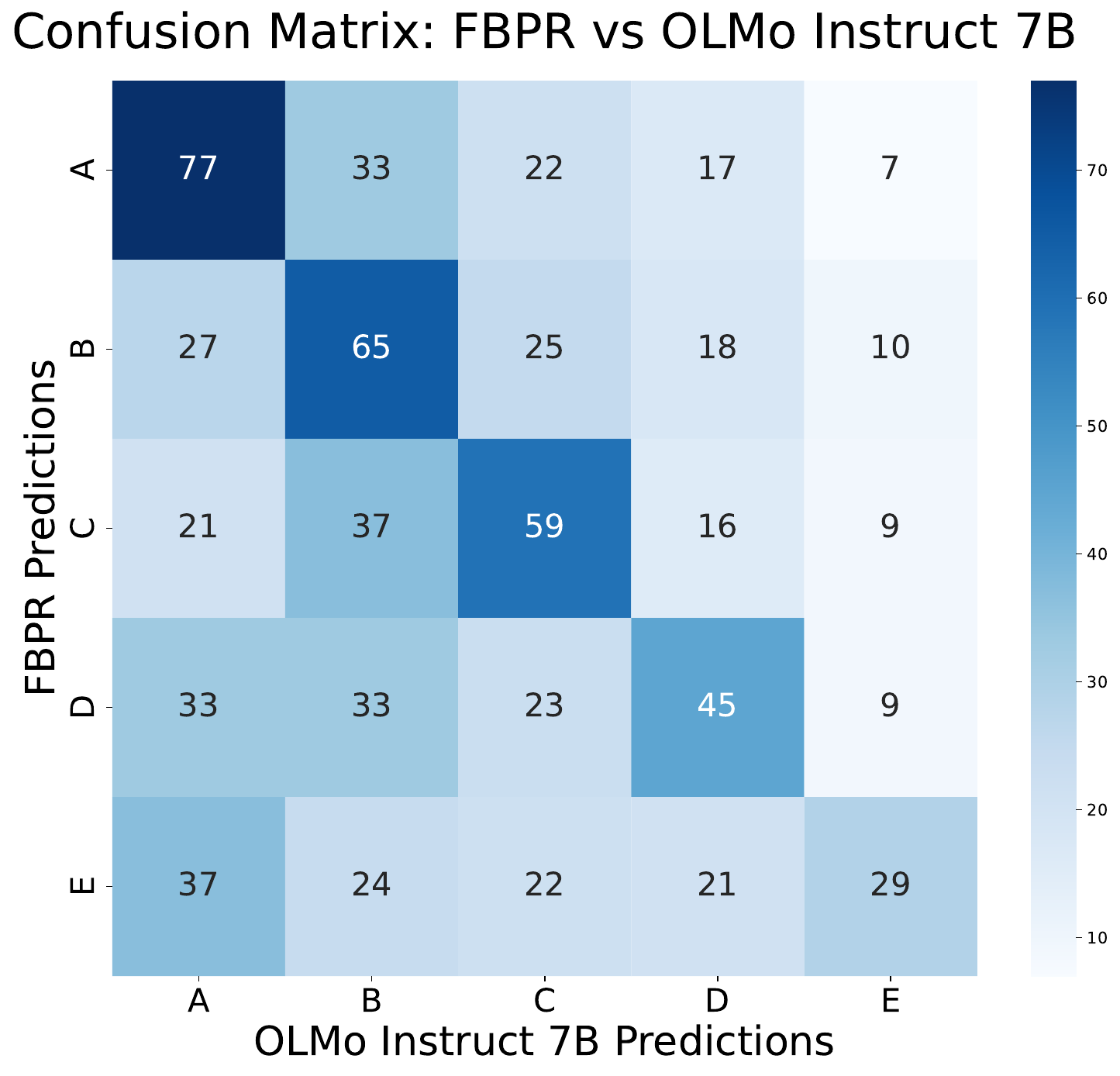}
    \caption{Confusion matrix between \modelNameAbbrev{} on Dolma and OLMo Instruct 7B on \textsc{MedQA} diagnosis subset.}
    \label{fig:olmo_confusion_matrix}
\end{figure}

\subsection{Certainty results}

\begin{figure}[htbp]
    \centering
    \includegraphics[width=0.8\linewidth]{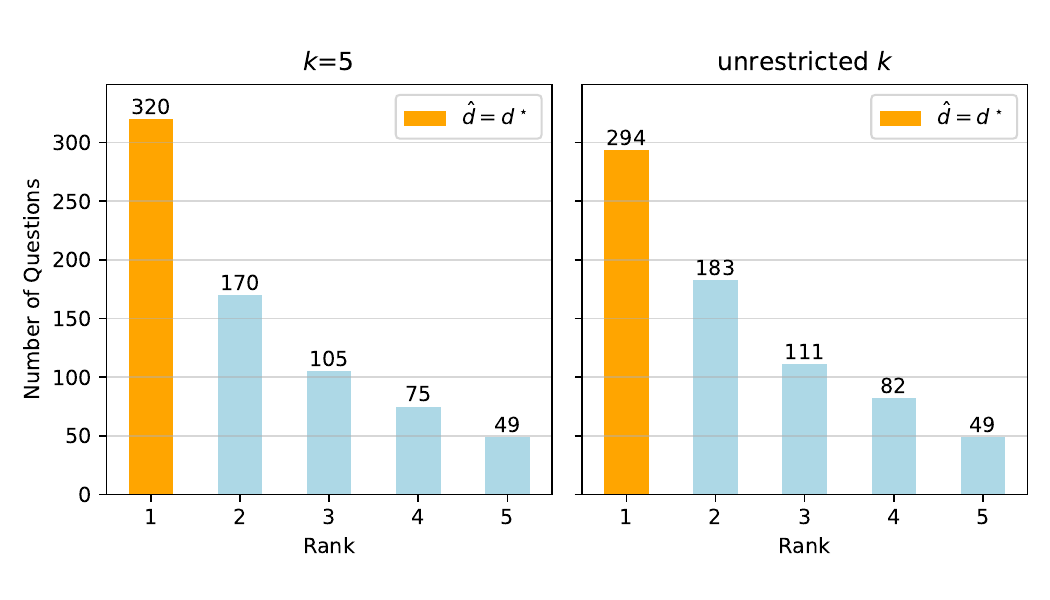}
    \caption{Distribution of the rank of the correct diagnosis of each question by RedPajama-based \modelNameAbbrev{} with $k=5$ or unrestricted $k$.}
    \label{fig:rpj_ignoreNegated_correctOption_ranks}
\end{figure}

\newpage
\subsection{\modelNameAbbrev{} pipeline examples}
\label{app:examples}
\begin{figure*}[htbp]
    \centering
    \includegraphics[width=\linewidth]{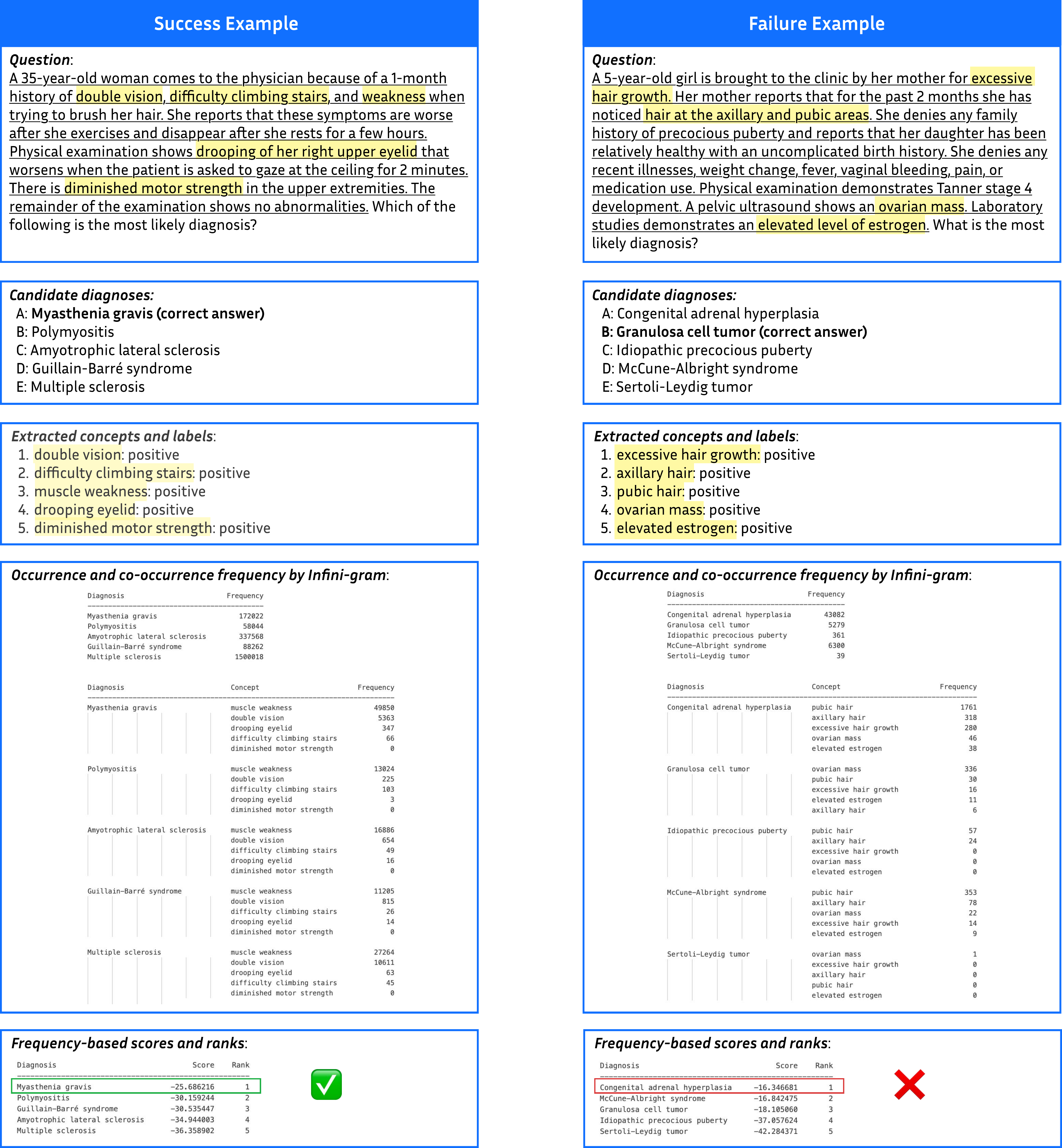}
    \caption{A success example (left) and a failure example (right) of \modelNameAbbrev{} on \textsc{MedQA} diagnosis subset. The diagnosis occurrence and (diagnosis, concept) co-occurrence frequencies are retrieved by \textit{Infini-gram} from \texttt{v4\_rpj\_llama\_s4} index, i.e., from the RedPajama corpus. By \textit{Infini-gram} design, some of these frequencies are estimates.}
    \label{fig:examples}
\end{figure*}

A success ($\hat{d} = d^\star$) and a failure ($\hat{d} \neq d^\star$) example of \modelNameAbbrev{} are shown in Figure~\ref{fig:examples}.

\clearpage
\section{Additional statistical analysis}
\label{app:stats}
We further quantify uncertainty for results reported in Table~\ref{tab:methods_accuracy} with 95\% bootstrap confidence intervals (10,000 paired resamples over the 719 \textsc{MedQA} questions) and test paired differences with McNemar's exact test (Table~\ref{tab:stats_ci}).

\begin{table}[htbp]
\centering
\begin{tabular}{lcc}
\toprule
Method & Accuracy & 95\% CI \\
\midrule
\modelNameAbbrev{} on Dolma & 46.7\% & [43.1, 50.3] \\
OLMo Instruct 7B & 44.1\% & [40.5, 47.7] \\
Hybrid (OLMo + Dolma) & 47.3\% & [43.7, 50.9] \\
\midrule
\modelNameAbbrev{} on RedPajama & 44.5\% & [40.9, 48.1] \\
LLaMA 65B & 47.0\% & [43.4, 50.6] \\
Hybrid (LLaMA 65B + RedPajama) & 50.8\% & [47.0, 54.4] \\
\bottomrule
\end{tabular}
\caption{Accuracy on the \textsc{MedQA} diagnosis subset ($n=719$) with 95\% bootstrap confidence intervals.}
\label{tab:stats_ci}
\end{table}

The accuracy difference between \modelNameAbbrev{} and its corpus-aligned LLM is not statistically significant for either pair (McNemar $p=0.30$ for \modelNameAbbrev{} on Dolma vs.\ OLMo Instruct 7B; $p=0.33$ for \modelNameAbbrev{} on RedPajama vs.\ LLaMA 65B). The two methods show only weak agreement beyond chance (Cohen's $\kappa=0.23$ [0.18, 0.27] for \modelNameAbbrev{} on Dolma vs.\ OLMo Instruct 7B and $0.19$ [0.15, 0.23] for \modelNameAbbrev{} on RedPajama vs.\ LLaMA 65B).

We apply Holm correction across the four pairwise tests comparing the hybrid method with \modelNameAbbrev{} and the LLM across the two corpora. For LLaMA 65B + \modelNameAbbrev{} on RedPajama, the hybrid improves significantly over both \modelNameAbbrev{} ($+6.3$ points, 95\% CI [2.5, 10.2], Holm-adjusted $p=.007$) and the LLM ($+3.8$ points, [1.0, 6.5], $p=.032$). For OLMo + \modelNameAbbrev{} on Dolma, the gains are not significant after correction ($+3.2$ points over the LLM, [0.4, 6.0], $p=.059$; $+0.6$ points over \modelNameAbbrev{}, [$-3.3$, 4.3], $p=.83$).